\documentclass[sigconf]{acmart}

\usepackage{tikz}
\usetikzlibrary{shapes,positioning}
\usepackage{dsfont}
\AtBeginDocument{%
  \providecommand\BibTeX{{%
    \normalfont B\kern-0.5em{\scshape i\kern-0.25em b}\kern-0.8em\TeX}}}

\setcopyright{acmcopyright}
\copyrightyear{2023}
\acmYear{2023}
\acmDOI{XXXXXXX.XXXXXXX}

\acmConference[KDD '23]{29th ACM
SIGKDD Conference on Knowledge Discovery and Data Mining}{August 06--10,
  2023}{Long Beach, CA}
\begin{document}

\title{Development of an End-to-end Machine Learning System with Application to In-app Purchases}

\author{Dionysios Varelas}
\affiliation{%
  \institution{Activision Blizzard King}
  \city{London}
  \country{UK}}
\email{dionysis.varelas@king.com}

\author{Elena Bonan}
\affiliation{%
  \institution{Activision Blizzard King}
  \city{Barcelona}
  \country{Spain}}
\email{elena.bonan@king.com}

\author{Lewis Anderson}
\affiliation{%
  \institution{Activision Blizzard King}
  \city{London}
  \country{UK}}
\email{lewis.anderson@king.com}

\author{Anders Englesson}
\affiliation{%
  \institution{Activision Blizzard King}
  \city{Stockholm}
  \country{Sweden}}
\email{anders.englesson@king.com}

\author{Christoffer Åhrling}
\affiliation{%
  \institution{Activision Blizzard King}
  \city{Stockholm}
  \country{Sweden}}
\email{christoffer.ahrling@king.com}

\author{Adrian Chmielewski-Anders}
\affiliation{%
  \institution{Activision Blizzard King}
  \city{Barcelona}
  \country{Spain}}
\email{adrian.chmielewskianders@king.com}

\renewcommand{\shortauthors}{Varelas, Bonan, et al.}

\begin{abstract}
  Machine learning (ML) systems have become vital in the mobile gaming industry. Companies like King have been using them in production to optimize various parts of the gaming experience. One important area is in-app purchases: purchases made in the game by players in order to enhance and customize their gameplay experience. In this work we describe how we developed an ML system in order to predict when a player is expected to make their next in-app purchase. These predictions are used to present offers to players. We briefly describe the problem definition, modeling approach and results and then, in considerable detail, outline the end-to-end ML system. We conclude with a reflection on challenges encountered and plans for future work.  
\end{abstract}

\keywords{Machine Learning, Machine Learning Platform, Mobile Gaming, Personalization, In-app Purchases}
\maketitle

\section{Introduction} \label{introduction}
Players of King games (e.g. Candy Crush Saga, Candy Crush Soda Saga), progress through a map of levels of increasing difficulty by solving match-3 puzzles. The speed of progression depends on the player's skill level which is a combination of deciding on the right move to make, the right boosters to use and the right time to use them. For example, a skilled player would know exactly when to use a chocolate bomb to overcome a really challenging game board. Not all players need extra boosters and those who need extra do not purchase them with the same frequency. Indeed, purchase behavior varies a lot from player to player and depends on multiple characteristics related to both long-term and short-term player behavior.

Machine learning (ML) allows us to combine a large number of behavioral and game-related features to predict when a player will make their next in-app purchase. This insight can inform our personalization engine to offer unique gameplay experiences to every player. In section \ref{use cases} we will expand on the use case for personalized offers.

\section{Offer personalization} \label{use cases}
Historically, personalized experiences at King were achieved using rule based approaches that were based on large player segments. Customizing the in-game store with individualized offers means that each player can purchase the boosters that they like.
To display an offer in the game, we have to answer the following questions: 
\begin{itemize}
\item Where is the offer shown (e.g. in the in-game store and/or with a pop-up)?
\item Which items are in the offer (e.g. boosters and/or gold bars)?
\item When is the offer available to the player (e.g during weekdays or weekend)?
\end{itemize}
Our motivation for this work was to answer the last question in a personalized way. Indeed, to decide on which day we show the offer to the player, we have utilized an ML model which predicts when the next in-app purchase is most likely to happen. To answer the other two questions about offer location and contents, we have defined and tested several configurations based on the lessons gathered from previous work in this area.
In section \ref{sec:previous_offers}, we describe how offer personalization was previously implemented in one of our games.

\subsection{Data-driven timing for personalized offers} \label{sec:previous_offers}
Players in our games exhibit vastly different patterns of engagement and purchase behavior. Such irregularity in their engagement with the store introduces the need to present offers at different time intervals. The goal is to learn the best time intervals from player's purchasing patterns.
In previous iterations, we had employed a fixed time interval for every player, meaning that we would wait \(n\) number of days before showing the offer.
Following this, we analyzed the purchase behavior of players in the game and observed three large segments based on their past transactions. Looking at the distribution of the number of days between purchases for each segment, we decided some thresholds to test based on different percentiles. 
Regarding the location of the offer, we decided to use in-game pop-ups because they are very flexible and quick to implement. The idea was first to test the reaction of the players to the offer using rule based iterations, then move to a personalized approach using machine learning. Since these first iterations were successful, we started to work on the model described in section \ref{models}. 

\begin{figure}[ht]
    \centering
    \includegraphics[width=0.3\textwidth, height=0.4\textheight]{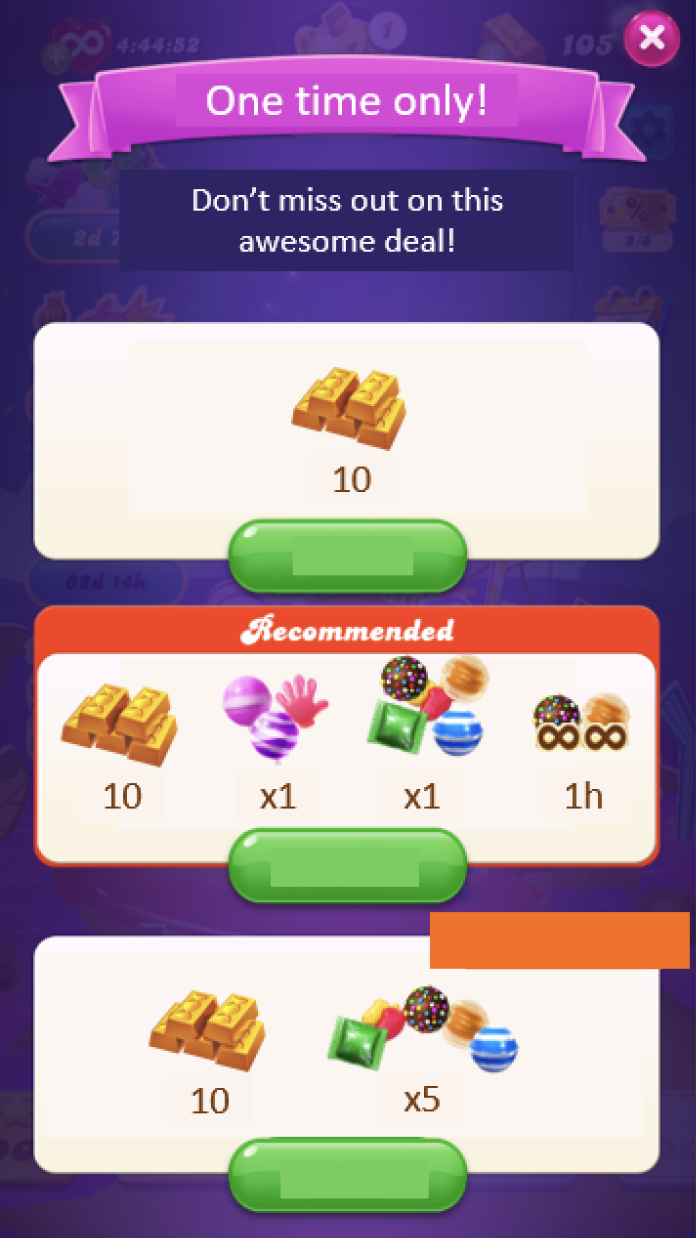}
    \caption{Discounted offer}
    \label{fig:soda-discount}
\end{figure}

\section{Problem Definition}

The target of this model is to identify for each player the probability of making a purchase any day within the next \(N\) days. Essentially, this requires the estimation of the cumulative distribution function (CDF) for each player over the following \(N\) days of interest. According to this, a player is eligible to receive an offer if they have not made a purchase until the day that corresponds to the median of the CDF. 

Formally, let $D={d_1,...,d_N}$ denote the day a player will make a purchase and $X_i={x_{i_{1}},...,x_{i_{K}}}$ the features considered for player $i$. Further, let \(p^{(i)}\) denote player \(i\) 

. Our problem can be formulated as a conditional probability distribution estimation of the following expression
\[
    P \big( p^{(i)} \text{ purchases on } d_n \mid X_{i} \big) 
\]
After estimating the above quantity we derive the CDF by summing up the individual probabilities for each day. 

\[
    P 
    \big( 
        p^{(i)} \text{ purchases in } 
        d_1, \dots, d_N
        \mid X_i
    \big) 
    = \\
    \sum_{n=1}^N P
    \big(
        p^{(i)} \text{purchases on } d_n
        \mid X_{i} 
    \big)
\]

\section{Experimentation}

Offline experimentation is the first step before moving to productionize an ML model. As part of this process, ML practitioners experiment in two key areas: feature engineering and modeling. The goal is to build a model that improves both quantitative metrics, such as predictive accuracy, and qualitative metrics as defined by business experts.

\subsection{Feature engineering}

The feature engineering process consists of extensive feature exploration and transformations. There are three categories of features to consider:

\begin{itemize}
  \item Play behavior (e.g. number of games played in a day)
  \item Purchase behavior (e.g. number of purchases in a day)
  \item Contextual features (e.g. day of week)
\end{itemize}

Typically, ML practitioners craft new features, using the types above as a basis, by aggregating over time windows, computing ratios and other custom transformations leading to more than 200 features to try out during modeling \cite{Haldar_2019}.

\subsection{Modeling}

Usually with a managed Jupyter notebook service, ML practitioners experiment with model configurations in order to optimize offline metrics. A good model for purchase day prediction should be able to predict accurately the day players are going to make their next purchase. In the case the model does not predict the exact day accurately we want it to be as close as possible to the truth. Based on this, we used both classification and regression metrics to account for the ordinal nature of the value we are trying to predict. In addition to those, the candidate models are evaluated against qualitative metrics as defined by the use case needs.

A model configuration consists of:
\begin{itemize}
  \item Model family (e.g. neural network, decision tree)
  \item Model hyperparameters (e.g. number of layers, number of nodes)
  \item Feature set (e.g. a subset of the engineered features from the previous step, volume of training data to use)
\end{itemize}

 In section \ref{models}, we talk about the model evolution before settling on the one we decided to test in an online environment.

\section{Models} \label{models}
Model development is an iterative process, starting with a really simple model and then working incrementally by tweaking small parts of the configuration in order to identify the changes that led to significant boost in offline metrics.

\subsection{Popularity baseline}

The first approach we tried was based on purchase day popularity for three segments of players. The segments were defined using purchase frequency. For each segment we calculated the most frequent day on which each player made their next purchase (e.g. next day, after two days etc.). Let \(S = S_1 \cup S_2 \cup \dots \cup S_M\) be the partition of players, then 

\[
    P \big( 
        p^{(i)} \text{ purchases on } d_n
    \big)
    = 
    \frac
        {
            \text{count(} 
                \{
                    p \text{ purchases on } d_n \mid p \in S_m
                \}
            \text{)}
        }
        {
            \text{count(}S_m\text{)}
        }
\]

Here, \(i\) ranges from 1 to $N$, where $N$ is maximum day ahead we were trying to predict and $M=3$. In this case, player $p^{(i)}$ belongs to segment $S_m$.

This was a simple model to build. After we calculated the quantity above for each $(d_n, S_m)$ combination we used the maximum probability value for each segment as the predicted purchase day. This was a quite powerful baseline given the nature of the problem we were trying to solve.

\subsection{Tabular neural network}

Improving the offline metrics against the popularity baseline is not a trivial task. We rely on a large dataset of player features which span over 1 year. Additionally, we use a custom neural network architecture which captures the interactions between those features and model the problem by taking into account the ordinal nature of the label. The model combines all features from the feature engineering process (play behavior, purchase behavior, contextual features), learns embeddings for all the categorical entities, and concatenates everything into a dense layer which is followed by two hidden layers leading to a softmax output layer. Specifically, the model configuration is the following:

\begin{itemize}
  \item All the continuous features are propagated to the concatenation layer.
  \item For all the categorical features we build embedding layers.
  \item The learned embeddings are propagated to the concatenation layer.
  \item Dense hidden layer 1 after the concatenation layer with relu activation of size 64.
  \item Dense hidden layer 2 after hidden 1 with relu activation of size 32.
  \item Softmax output layer of size 16.
\end{itemize}

The softmax output is fed into a cross entropy (CE) loss and the argmax of the 16 probabilities of the softmax is fed into mean squared loss (MSE). We use two hyperparameters, $a$ and $b$, which we tune using cross validation.

\begin{equation*}
    \label{eq:loss}
    \mathrm{loss} = a\cdot \mathrm{CE} + b\cdot \mathrm{MSE}
\end{equation*}

\subsection{ContentRNN}

The tabular NN is good at capturing the long term behavior and interest of players but lacks the ability of capturing short term signals.

For that purpose we enhance the existing architecture by incorporating a time series feature able to capture this information. We introduced ContentRNN, an RNN model \cite{bahdanau2016neural} augmented with content based features. In addition to this, we add a calibration layer \cite{Calib} in order to get better estimates of the probabilities coming out of the softmax. After training the model end to end, we run a second pass having all the layers frozen apart from the calibration layer.

\begin{figure}[ht]
    \centering
    \includegraphics[width=0.5\textwidth]{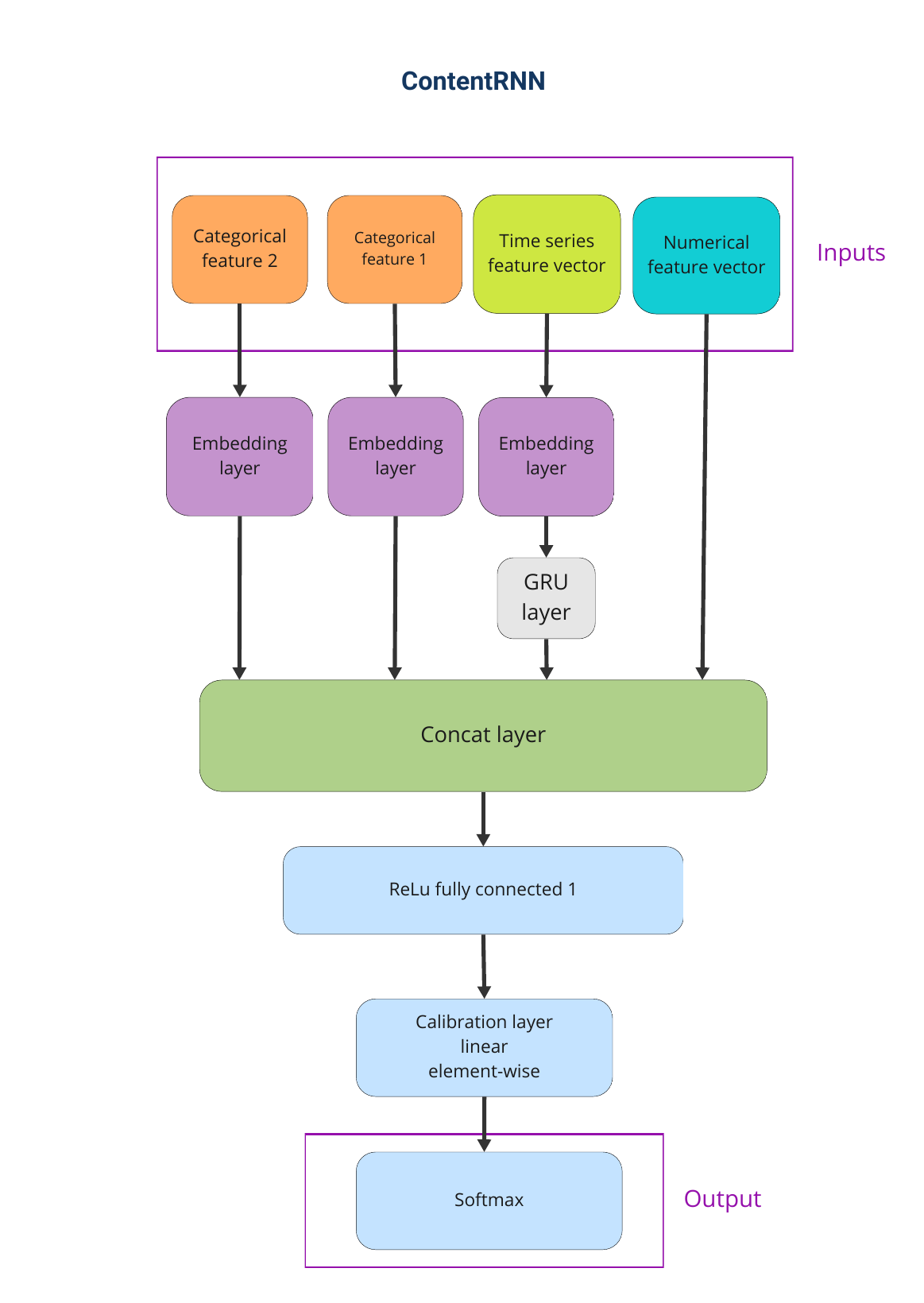}
    \caption{ContentRNN Architecture}
    \label{fig:contentrnn}
\end{figure}

This architecture and training strategy led to improvement of the offline metrics.

\subsection{Offline results}

\begin{figure}
    \begin{tabular}{|r||c|c|c|c|}
        \hline
        \textbf{Model} & \textbf{MAE} & \textbf{Accuracy} & \textbf{Precision} & \textbf{F1} \\
        \hline
        Popularity baseline & 3.84 & 0.39 & 0.18 & 0.25 \\
        \hline
        ContentRNN & 3.36 & 0.41 & 0.27 & 0.29 \\
        \hline
    \end{tabular}
    \caption{Offline results of both models on a proprietary test dataset.}
    \label{fig:results}
\end{figure}

Figure \ref{fig:results} details the offline results of the two models evaluated against our proprietary dataset. \\

\subsection{Online results}

Improving the experience of our players is our main driver and evaluating our approach online is the only way to identify if our model has positive impact in player's gaming experience. It is common that improvements in offline metrics do not translate to improved experience. In order to measure the impact of ContentRNN we ran A/B tests. We evaluated the model against key business metrics which are considered sensitive and cannot be shared. In the most recent A/B test ContentRNN was evaluated against the popularity baseline which already proved to have positive business impact. Against the popularity baseline we observed 20\% increase on one of the key business metrics.

\section{ML Platform}

The process of building and productionizing ML models within King has been in place for several years; however, different teams use a wide variety of methods, libraries and environments to deliver their work. The lack of a single unified workflow has led to huge inefficiencies and duplication of effort across the company, creating an environment where ML is considered difficult to do at scale.

To better enable the experimentation and deployment of machine learning models, King has built a platform which assists and automates parts of the ML workflow \cite{scalable_ml_systems}. Using a self-service approach, a number of modular components and tools which aid the modeling workflow are being offered to machine learning practitioners and aligned under a common system in similar fashion to previously described ML systems \cite{looper, uber_ml, tfx, managing_ml_vartak}.

\subsection{Data sources}\label{subsec:datasources}

King's data warehouse, which is centred around analytics workloads, 
is now starting to cater to a new surge of ML use cases, where features are developed and shared for models such as the in-app purchase prediction model.

\subsubsection{Feature engineering platform}

The model was developed using data sourced from the data warehouse. During experimentation, numerous features were crafted that reflected both behavioral, demographic and contextual attributes for each player.

As a part of the push to a production ready system, the data preparation was moved into a prototype feature engineering platform (FEP).

This platform serves several purposes:
\begin{itemize}
    \item Generate the full player feature history.
    \item Schedule and execute the daily jobs for latest feature generation.
    \item Schedule and execute the daily labelling jobs.
    \item Provide an API for serving feature data into different formats.
    \item Notify downstream systems and processes of feature and label readiness.
\end{itemize}

\subsubsection{Feature Orchestration}

BigQuery SQL was used to generate the latest feature values and append them to the FEP storage table. This table contains not only the latest player-feature values, but an entire history of previous values. The FEP contains a module which programatically generates data validation tests from a small configuration file. Each day, hundreds of data unit tests are executed to ensure that the features are calculated as expected, giving confidence to ML practitioners that the data used for training and inference is correct.

The scheduling and orchestration of the jobs in the Feature Engineering Platform are done using Airflow, where several different DAGS are responsible for components of the workflow.

\subsubsection{Dataset serving}

Extracting batch sets of data from the FEP is needed for re-training and inference. The feature serving process is executed on demand, from the ML Platform re-training and batch inference components.

A pipeline written using the PySpark \cite{spark} processing framework serves data to these downstream processes. This data processing pipeline has several steps:
\begin{itemize}
    \item Read the data from the FEP storage table.
    \item Limit the number of samples processed (on request).
    \item Cast data types.
    \item Filter out certain features (on request).
    \item Calculate feature statistics.
    \item Write the data to the specified format (e.g. TFRecord, JSONL).
    \item Write the feature statistics to a file alongside the data (e.g. mean, stddev).
\end{itemize}

This Pyspark pipeline executes in a serverless environment in Google Cloud Platform, through the Dataproc batch service. The initiation of this job is exposed to certain users through HTTP Cloud Functions, enabling RESTful requests but with the in-built benefit of GCP Identity and Access Management (IAM) restrictions.

A Python library was developed to work with these cloud functions, wrapping the requests in a small installable Python client that allows users and processes to easily trigger the pipeline with custom parameters.

\subsubsection{Readiness Notifications}

Using our in-house data notification service, the FEP shares alerts to all data consuming users and systems to signify when any job has completed. Usage of the notification service ensures that feature data for all parts of the ML workflow is only consumed once it's made available, with zero wait time.

\begin{figure}
    \centering
    \includegraphics[height=0.55\textheight]{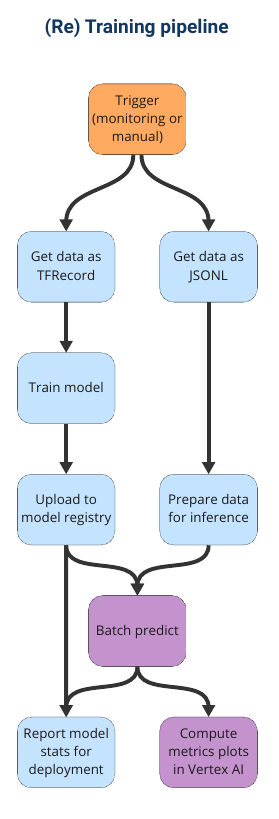}
    \caption{Structure of a (re) training pipeline. Purple colored elements represent standard operations in Vertex AI.}
    \label{fig:training_pipeline}
\end{figure}

\subsection{Training}\label{training}

Experimenting with different model architectures and
features is orchestrated with an internal tool called KingML. KingML allows data scientists at King to orchestrate training pipelines which run on the Vertex AI platform \cite{Bisong2019}. By abstracting the concept of pipelines KingML offers a minimal setup for data scientists to run their training code on their local computer, a Jupyter notebook, or remotely in Vertex AI. When running remotely, artifacts and metrics are all saved to Vertex AI so easy comparisons can be made between runs. After experimentation, these training pipelines can then be promoted to a production environment and used for retraining purposes with no additional configuration.

The training pipeline is made up of the following steps: data ingestion, data transformation, training, model upload, model evaluation and user notification.

\subsubsection{Data ingestion}

The FEP API will materialize datasets on demand. For training, we create a dataset with 16 months worth of data. This dataset is divided into three subsets. The training set is made up of 12 months of data while the the validation set and test set both consist of 2 months of data.

We use TFRecords as the data format and leverage the Tensorflow Dataset API for managing the data ingestion pipeline during training.

\subsubsection{Data transformation}

While the FEP takes care of imputation of missing data the training pipeline is responsible for the data transformation. The transformations that are applied in this pipeline are the normalization of features and capping the values of certain features.

These transformation steps are then later stored in the produced model artifact to prevent training-serving skew.

\subsubsection{Model Training}
The training code is written in Python which KingML can run locally and in the cloud. KingML builds and pushes a Docker image that encapsulates the training logic to run in Vertex AI. In this setup, Google Cloud Storage (GCS) is mounted as a local file system which enables the training logic to read data without having to check where it may be. The output of the training step is a model artifact.

\subsubsection{Model upload}
After a model is trained, a new model version is created under the existing model in the Vertex AI model registry.

\subsubsection{Model evaluation}
A batch predict step is done in order to obtain predictions on the validation set. We use these predictions in conjunction with the ground truth labels and Vertex AI to compute precision, recall, ROC curve, confusion matrix, accuracy and other metrics. This is tied to the model version in the model registry for easy viewing. 

\subsubsection{User notification}
Upon completion of the training pipeline, emails are sent to the data scientists responsible for the model informing them the pipeline has finished along with links to view the result and metrics. 

When a model has been promoted to a production environment, predictions from inference are being used in live games. One of the steps in this promotion is the configuration of a monitoring policy which will trigger the associated retraining pipeline. Configuration is done via an internal service called Schedulai, which manages ML training pipelines and receives webhook notifications when monitoring incidents occur and maps these incidents to models and retraining pipelines.

\begin{figure*}[h]
    \centering
    \includegraphics[width=1\textwidth]{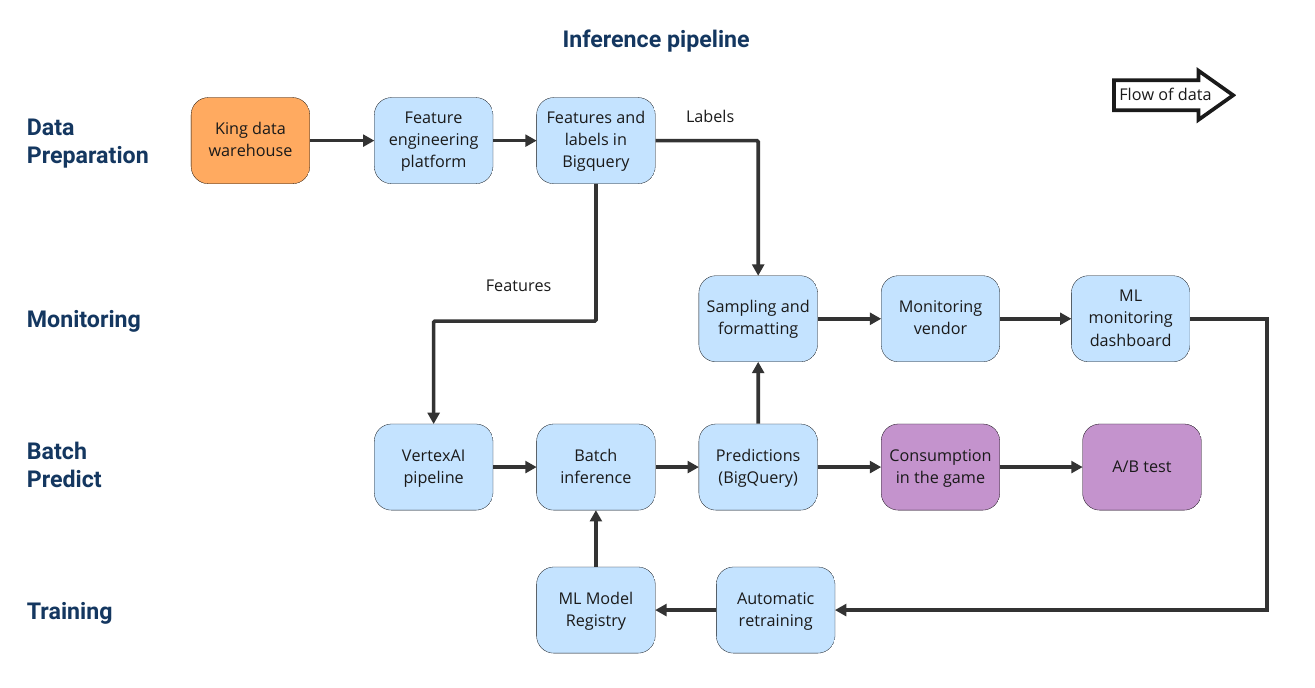}
    \caption{Inference pipeline}
    \label{fig:infpipeline}
\end{figure*}
\subsection{Deployment}
The deployment component of the platform is responsible for providing the daily predictions to the game using data from the FEP and the model produced by the training process. Its goals are to:
\begin{itemize}
\item Ensure that the inference pipeline performs the same transformations as in training
\item Guarantee scalability to handle hundreds of millions of predictions per model
\item Make predictions available to our personalization engine
\item Send a notification as soon as the predictions are ready to be consumed
\end{itemize}

\subsubsection{Pipeline trigger}
As soon as the daily data is available in the FEP, a message is sent to a pub/sub topic using an internal notification system. Then a pub/sub subscription to the topic forwards the message to a GCP Cloud Run App which triggers a GCP Vertex AI pipeline.
\subsubsection{Batch predict pipeline}
We use Vertex AI for batch prediction pipelines because it allows us to access easily all the GCP services and to manage Kubeflow \cite{Bisong2019} pipelines. For this reason, we have developed internal tools to speed up the implementation and orchestration of Vertex AI pipelines. 
\subsubsection{Pipeline components}
We utilize Dataproc's serverless components, as they offer the benefits of serverless services and the capabilities of PySpark. By leveraging PySpark, we can efficiently handle large volumes of data and replicate data manipulations conducted during experimentation using Python.
\subsubsection{Batch predictions}
To get the predictions, first the daily features (see \S \ref{subsec:datasources}) are loaded and normalized using statistics from training (see \S \ref{training}). Then the model provided by the training component (see \S \ref{training}) is used to predict in batches.

\subsubsection{Prediction Storage}
The predictions are pushed to BigQuery. We use an internal notification system to notify the personalization engine that the predictions are ready to be consumed. 
\subsection{Monitoring}
Model monitoring is an important part of reliable and production-level machine learning systems \cite{monitoring_score}. Indeed, there are a lot of factors that can threaten the expected performance of a ML model. For example:
\begin{itemize}
    \item Difference between the training and inference processes (training/serving skew)
    \item Change in the feature distribution (covariate shift)
    \item Change in the relation between features and label (concept shift)
\end{itemize}
  The system continuously monitors input features, predictions and labels to be able to quickly detect these issues and allow ML practitioners to apply the right solution. Some possible solutions include: retraining the model, fixing  a bug in the data pipeline or redesigning features to be more resilient against distribution shift. \cite{diethe2019continual}
After careful consideration of whether to build or buy a solution, we have opted to bring on board a third-party vendor to accelerate the adoption of ML monitoring. The tool comes with: 
\begin{itemize}
    \item Daily metrics related to the distribution of the input features
    \item Performance metrics (e.g. accuracy)
    \item Analysis by custom segments and time frames
    \item ML monitoring policies  \& Alerts
    \item Slack \& email integration
    
\end{itemize}
The third party monitoring tool easily integrates into the platform thanks to its API and SDK.
\subsubsection{Monitoring of predictions and features}
After the predictions are stored in BigQuery, we select a sample of input features with the corresponding predictions and push them to the monitoring platform. It is important to ensure that the sample size is adequate to represent the population accurately. In this particular case, the sample size was calculated to ensure that the statistical properties of both the features and predictions distributions were not impacted. The approach to determining the sample size was to select the smallest number of data points that would result in the least possible distributional distance between the original features and the sampled features.
\subsubsection{Monitoring of labels} \label{labels}
In order to enable performance monitoring (e.g. calculate metrics like accuracy, MAE, etc.), we need to observe the actions taken by the players (i.e. collect the labels).
The labels are being gathered with delay as we need to wait for $N$ days in order to observe if a purchase has been made or not. Every day we push the labels to the monitoring platform which are matched to the corresponding prediction.
\subsubsection{Monitoring policies}
Monitoring policies include a daily scan of all the data logged in our monitoring platform to detect any anomalies. If a monitoring policy is found to be violated, an incident is generated and the model's owners are promptly notified via email and Slack. Specifically, we focus on detecting shifts in important features, shifts in predictions, changes in performance, and any missed daily predictions. 
\subsection{Infrastructure}

We use Terraform and Git to manage infrastructure as code, ensuring consistency and reproducibility across different environments. By using Terraform, we are able to easily track changes over time and set up infrastructure for new use cases. Access control and permissions for Google Cloud resources are managed through IAM, where the principle of least privilege is followed to provide ML platform users with access only to the resources and actions required for their job. Code deployment is managed via Cloud Run, which deploys stateless HTTP containers that automatically scale up and down based on traffic. Docker containers are built via Cloud Build CI and pushed to Container Registry in GCP, which makes tracking versions easy.

In terms of storage, we use BigQuery to store preprocessed data and model predictions. GCS buckets are also utilized for intermediate storage for some processed data, such as DataProc jobs, before being stored in BigQuery tables. Access control to the buckets and BigQuery was easily set up through IAM with Terraform. Finally, a Google Virtual Private Cloud (VPC) was set up to access data and services from King's on-premise network, VertexAI pipelines, DataProc jobs, and services. With these tools and configurations in place, we are able to effectively manage infrastructure and data, while ensuring secure access to resources and services.

\section{Challenges}
Deploying and maintaining a large scale ML system comes with unique challenges. Our ML platform is in a constant state of evolution to overcome new challenges as they arise. In this section we present initial challenges and the lessons learned from those while building the platform.

\subsection{Retraining policy}

Finding the right time to retrain a model can proved to be difficult. Being late can reduce the model's performance due to shifts in data distributions. Being early leads to unnecessary use of resources that increase operational costs without any benefit. For that reason we configured our retraining policies in a way that balances between the two. We treat drops in performance (e.g. accuracy) as severe events and we trigger retraining almost instantly. However,  performance readings can come with delay (see \S \ref{labels}). For this we fallback to prediction distribution monitoring. We use a low sensitivity policy which monitors the distribution of the model output and when a big shift is observed we trigger retraining. A big shift is defined when a change greater than 3 standard deviations is observed.

\subsection{Training serving skew} Before the existence of the ML platform, a lack of a standardized approaches to building and maintaining data pipelines led to discrepancies between the model outputs during experimentation and in production. This became more apparent with the introduction of ML monitoring which showed those differences in feature and prediction distributions. Sometimes the discrepancies were small, like minor schema differences (float32 vs float64). Other times they were major like different pre-processing applied during training and serving. The FEP has solved this by standardizing the way features are being generated both during training and serving.

\subsection{Communication} One of the goals of the ML platform is to have a coherent and unified product that hides away all the different integrations within the platform, giving a harmonized experience to the ML practitioner. However, having a modular ML platform comes with a communication overhead and achieving our goal became a big challenge. One key learning was the need for contracts between each component early on and to have them be reviewed often as more components are added. With time and after multiple iterations those contracts will become more strict as more ML systems are developed and more general components are created.

\subsection{Production-first mindset} ML practitioners tend to be experts in understanding the use case domain. This skill makes it easy for them to experiment with data and models offline in order to solve business problems. However, there is a gap between offline experimentation and online model serving. Some of the offline experiments might be hard to move to production. Main reasons for this include online data availability, ML framework used, prediction latency. In order to solve this, ML platform engineers advise ML practitioners on best practices and help them during feasibility studies. In the future, we plan to provide more formal training in ML system design.

\section{Future Work}
As the need for ML systems continues to grow, so does the need for better frameworks and technologies. We have identified several areas of focus for the future.

\subsection{Feature store}

The FEP that was tested by this model was a significant step towards the support of a data catalog of ML features that are crowd sourced and widely available. In the near future we hope to develop a company wide feature store from which pre-computed and commonly used features can be distributed \cite{featurestore_whatis}. The aim is to make the development of features for ML use cases significantly easier, more cost effective, and also to reduce the risk of mistakes through data leakage and inaccurate feature computation \cite{featurestore}.

\subsection{Deployment strategies}

Providing robust and reliable ML-driven products requires flexible deployment strategies. A simple deployment strategy would be to enable a new ML feature for every player. This in practice is dangerous as many unforeseen failures might appear in production. For this reason we are working on providing an easy way to shadow deploy models before going live. A shadow deployment records all the production data and the outputs of the machine learning model without resulting in an action. Players receive the default experience and not the one coming from the ML model. This combined with our ML monitoring capabilities will allow ML practitioners to get an understanding of what would happen if the model was live.

\subsection{From batch to real time predictions}

At present, the majority of the machine learning systems are deployed via batch inference. Being a able to serve more use cases in real time would allow for more sophisticated models and reduce the computation of prediction scores that are never used by our clients. We expect to see great benefits both in increasing the predictive power and usefulness of the models and reducing the operational costs by computing scores on demand only.

\subsection{Explainable AI (XAI)}

Explainability is a key driver of ML adoption in every industrial setting \cite{BARREDOARRIETA202082, xai_principles}. Being able to explain predictions coming out of ML models provides confidence to adopters as models stop being treated as black-box decision engines. The goal here is to be able to understand easily the relationship between input data and model outputs.

\section{Acknowledgments}
We thank Daniel Neiberg who brought up the business case and the Candy Crush Soda analytic team who helped in the implementation of the AB test. We thank Alex Nodet, Péter Németh, Daniel Lind, Martin Lundholm, Simon Westberg, Alvaro Garcia Carasso and the rest of the ML Platform team who collaborated in the productionization of this machine learning system. 

\bibliographystyle{ACM-Reference-Format}
\bibliography{references}

\end{document}